\documentclass[]{bytedance_seed}

\usepackage[toc,page,header]{appendix}
\usepackage{amsmath,amssymb}
\usepackage{enumitem}
\usepackage{needspace}
\usepackage{array}
\usepackage{tabularx}
\usepackage{colortbl}
\usepackage{longtable}
\definecolor{warnred}{HTML}{B44B4B}
\makeatletter
\def\fps@figure{t}
\def\fps@table{t}
\makeatother
\newcolumntype{Y}{>{\raggedright\arraybackslash}X}

\DeclareRobustCommand{\atrain}{{\bfseries\color{seedbg}A-Train}}
\DeclareRobustCommand{\atraintitle}{{\color{seedblue}A-Train}}
\DeclareRobustCommand{\titleul}[1]{{\color{seedblue}\renewcommand{\ULthickness}{1.4pt}\renewcommand{\ULdepth}{2.5pt}\uline{#1}}}
\title{\fontsize{18.5}{24}\selectfont\atraintitle{}: \titleul{A}gentic End-to-End Optimization\\[2pt]
for Efficient ML \titleul{Train}ing}

\author[1,2]{Hao~(Mark)~Chen}
\author[1,3]{Jiakun~Fan}
\author[1]{Dongsheng~Yang}
\author[1]{Neiwen~Ling}
\newcommand{\authornewrow}[2][]{\addtolist[#1]{#2}{\authorlist}{\authorformat}{\\}}
\authornewrow[1,4]{Yufa~Zhou}
\author[1,5]{Yuxuan~Xia}
\author[2]{Hongxiang~Fan}
\author[1]{Qing~Xiao}

\affiliation[1]{ByteDance}
\affiliation[2]{Imperial College London}
\affiliation[3]{Virginia Tech}
\affiliation[4]{Duke University}
\affiliation[5]{New York University}

\abstract{
General purpose ML training frameworks separate systems engineering from the learning algorithm, leaving large speedups unrealized while manual specialization across shifting models, devices, and datasets does not scale.
As AI systems accelerate research on their successors toward recursive self improvement (RSI), \textbf{training compute is becoming a primary bottleneck}.
We present \atrain{}, an autonomous agentic system where a single coding agent \textbf{co designs the learning algorithm and systems implementation} end to end to minimize time to a fixed quality target.
To balance verification cost against reuse, \atrain{} combines a minimal agent loop, \textbf{regularized search} with evaluator controlled recipes and hidden validation traces, and an \textbf{asynchronous, layered evaluator}.
Across six language and speech settings spanning datacenter GPUs, a consumer GPU, and a CPU, frozen trainers from the four fixed model, fixed device searches meet all main held out targets after recipe refitting, delivering \textbf{\mbox{3.77\textendash{}14.10{\boldmath$\times$}} speedups} over tuned PyTorch references, \textbf{3.51{\boldmath$\times$}} over tuned torchtitan on HBM3, and \textbf{up to 5.59{\boldmath$\times$}} over systems only search.
When hardware or models vary, agents embed \textbf{runtime self calibration and correctness guards} to transfer within validation coverage; more importantly, during search and discovery, they also invent methods including permutation sorted FP4 gradients, scale folding FP8 gradients, and zero weight compile twins.
These results establish \textbf{reusable, end to end workload specialization} as a scalable alternative to hand tuned training frameworks.
}

\begin{document}
\maketitle

\section{Introduction}
\label{sec:intro}

General purpose ML training frameworks~\citep{transformers,megatron,torchtitan,pytorch2} separate systems engineering from the learning algorithm to support many models, devices, and data shapes, leaving large speedups unrealized. As AI systems accelerate research on their successors toward recursive self improvement (RSI)~\citep{athanor,infrarsi,xia2026rrsi}, each cycle requires both intellectual discovery and compute to train the next generation. When more capable models speed up discovery, compute becomes the bottleneck, making further acceleration increasingly dependent on \textit{training efficiency}~\citep{ho2024algorithmic}.

\textbf{Workload specialization} offers a direct route to more efficient training. Tailoring both the learning algorithm and training code to a particular model and device can save substantial time~\citep{flashattention,deepseekv3,liger}, but the manual work of developing and testing a specialized trainer is costly to repeat~\citep{hooker2021hardware}. LLM coding agents have become capable enough on long horizon engineering and optimization tasks~\citep{metr_horizon,alphaevolve,vibeserve,autoresearch} to make it worth investigating whether they can autonomously optimize trainers for efficiency.

\begin{figure}[t]
\centering
\includegraphics[width=0.508\linewidth]{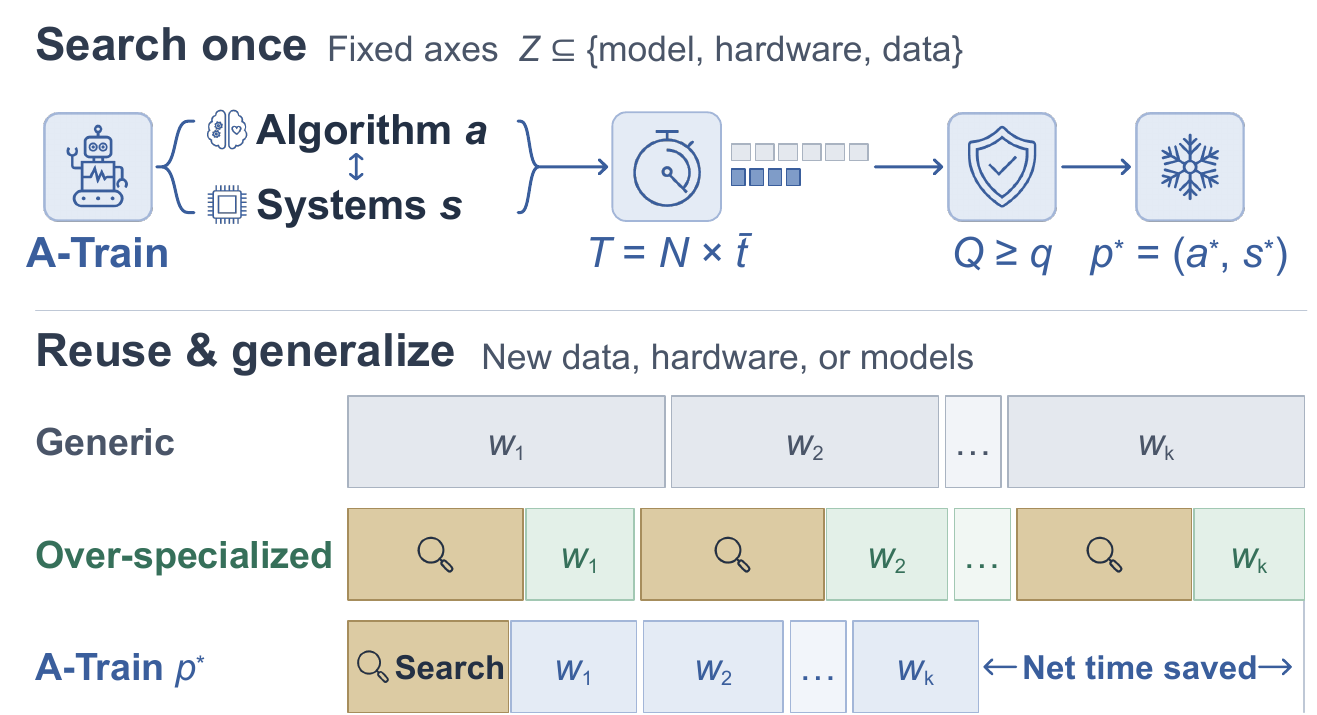}\hfill
\includegraphics[width=0.474\linewidth]{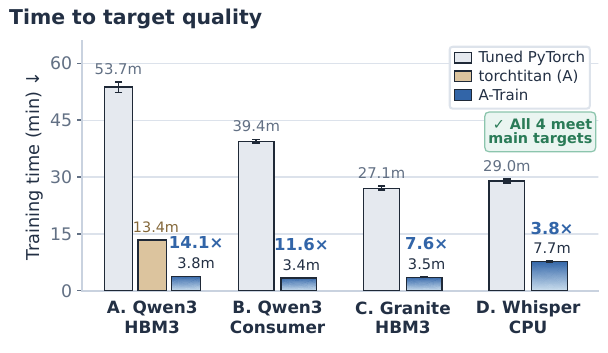}
\caption{\textbf{\atrain{} overview and results.} \textbf{(a)}~Co design of bespoke trainers $p^\star=(a^\star,s^\star)$, reused across datasets. \textbf{(b)}~Held out time to target quality in Scenarios~A--D.}
\label{fig:intro}
\end{figure}

However, building an autonomous agent system that reliably improves training efficiency via full-stack optimizations is challenging. A
trainer's learning algorithm and systems implementation jointly determine how long an update takes and how much progress it makes. Verifying a speedup can therefore require measuring \textit{end-to-end time to the same quality target}~\citep{algoperf,notrainnogain}. Those measurements also consume device time. For specialization to pay off, the trainer must save more device time across uses than the agent spends finding and validating it (\cref{fig:intro}a).

We model this tradeoff between \textit{the cost of search and the savings from reuse} (\cref{sec:framework}). The model motivates two design goals for \atrain{} (\cref{sec:system}). The first is to \textbf{balance the speedup on a particular workload against the range of workloads that can benefit from it}. The second is to \textbf{verify speed and correctness efficiently}. A minimal loop lets a single agent choose how to optimize the learning algorithm and its implementation, so the search can benefit from more capable agents. Regularization guides the search toward optimizations that exploit the target configuration and remain useful when other workload properties vary. A dedicated evaluator uses asynchronous execution and layered checks to verify training time and quality.

We test how far these optimizations transfer in six searches across language and speech tasks. In the four settings that test transfer to new datasets on the same model and device, \textit{all four trainers met their loss targets on the main held out tests}, achieving \(3.77\)--\(14.10\times\) speedups over independently tuned PyTorch references and \(3.51\times\) over tuned torchtitan on HBM3 (\cref{fig:intro}b), as well as up to \(5.59\times\) over systems only search (\cref{sec:results}). These results show that coding agents can improve training efficiency by jointly optimizing the learning algorithm and its implementation.

The paper makes three contributions.
\begin{itemize}[leftmargin=*,nosep]
  \item \textbf{An autonomous agentic system for trainer efficiency optimization.} We model when savings from reusing an optimized trainer repay the cost of verification. This principle guides \atrain{}, which jointly optimizes the learning algorithm and its implementation.
  \item \textbf{Trainer specialization that generalizes.} We show that \atrain{} can discover distinct optimizations for different devices and model architectures while retaining speedups on workloads unseen during search (\cref{sec:results,sec:adaptation}).
  \item \textbf{Agent behaviors and novel methods.} A single agent optimizes across the entire stack, tests and overturns its own evidence, and builds diverse local stress suites and runtime checks for unseen workloads (\cref{sec:discoveries}), while inventing specialized methods such as permutation sorted FP4 gradients, scale folding FP8 gradients, and zero weight compile twins (\cref{sec:novel}).
\end{itemize}

\newpage
\section{Agentic Search for Efficient Training}
\label{sec:framework}

\subsection{Problem Formulation}

\newcommand{\systemSymbol}[1][]{\textcolor[HTML]{73558C}{s#1}}
\newcommand{\algorithmSymbol}[1][]{\textcolor[HTML]{3F63AF}{a#1}}
\newcommand{\runtimeSymbol}{\textcolor[HTML]{A44620}{T}}
\newcommand{\qualitySymbol}{\textcolor[HTML]{91455F}{Q}}
\newcommand{\stepCount}[1][]{\textcolor[HTML]{4F5662}{N#1}}
\newcommand{\configSymbol}[1]{\textcolor[HTML]{006970}{#1}}

A trainer pairs an algorithm \(\algorithmSymbol\) with a system \(\systemSymbol\) that implements it. The agent specializes the trainer to a workload configuration \(\configSymbol{Z}\), which fixes a chosen subset of \(\{\text{model}, \text{hardware}, \text{data}\}\). For example, if the configuration fixes only the model and hardware, the trainer must remain reusable across datasets.

For a workload \(\configSymbol{Z}=\configSymbol{z}\), the agent minimizes training time subject to a checkpoint quality target~\citep{algoperf}.
\begin{equation}
  \min_{\algorithmSymbol,\systemSymbol}
    \;\runtimeSymbol_{\configSymbol{z}}(\algorithmSymbol,\systemSymbol)
  \qquad\text{subject to}\qquad
    \qualitySymbol_{\configSymbol{z}}(\algorithmSymbol,\systemSymbol)
    \geq q_{\configSymbol{z}} .
  \label{eq:shared-objective}
\end{equation}
Here \(\qualitySymbol_{\configSymbol{z}}(\algorithmSymbol,\systemSymbol)\) measures the candidate trainer's checkpoint quality. The target \(q_{\configSymbol{z}}\) is the reference quality on the same workload minus a tolerance calibrated across repeated reference runs.

Training time is \(\runtimeSymbol=\stepCount\,\bar t\), where \(\stepCount\) is the number of updates and \(\bar t\) is the mean time per update. Systems-only search fixes the reference algorithm \(\algorithmSymbol[_0]\) and update count \(\stepCount[_0]\), then minimizes \(\bar t\) over implementations that preserve the reference computation, denoted by \(\systemSymbol \simeq \systemSymbol[_0]\). Co-design varies the algorithm and system jointly, so it can reduce the number of updates needed to reach the target, the time per update, or both. Under the shared quality constraint in \cref{eq:shared-objective}, the two objectives are
\begin{align}
  \text{\textcolor[HTML]{666666}{Systems only}} \qquad
    &\min_{\systemSymbol\,\simeq\,\systemSymbol[_0]}
      \;\bar t(\algorithmSymbol[_0],\systemSymbol)
      \label{eq:system-objective}\\
  \text{\textcolor[HTML]{666666}{Co-design}} \qquad
    &\min_{\algorithmSymbol,\systemSymbol}
      \;\stepCount(\algorithmSymbol,\systemSymbol)\,
      \bar t(\algorithmSymbol,\systemSymbol).
      \label{eq:trainer-objective}
\end{align}
Every trainer feasible under systems-only search is also feasible under co-design, so the optimal end-to-end training times satisfy
\(\runtimeSymbol^\star_{\mathrm{joint}}\leq\runtimeSymbol^\star_{\mathrm{sys}}\).

\subsection{Cost-Benefit Modeling}

We model the net device time saved when an agent searches for a faster training program, then use the model to identify how to increase those savings. Let \(\mathcal U\) index future runs on which the selected program meets the workload's quality target, as defined for training in \cref{eq:shared-objective}. The net saving is
\begin{equation}
  G
  = \underbrace{\sum_{u\in\mathcal U}
      \bigl[\runtimeSymbol_{\mathrm{ref}}(u)-\runtimeSymbol_\star(u)\bigr]}
      _{\text{runtime saved across future runs}}
    - \underbrace{\sum_{k=0}^{K}\runtimeSymbol_k}
      _{\text{verification time}} .
  \label{eq:net-savings}
\end{equation}
Here \(\runtimeSymbol_{\mathrm{ref}}(u)\) and \(\runtimeSymbol_\star(u)\) are the runtimes of the reference and selected programs. Verification time includes measurements for both efficiency and correctness. The first term captures the tradeoff between \textbf{specialization} and \textbf{generalization}: specializing to a narrower class of workloads can lower \(\runtimeSymbol_\star(u)\) further on those workloads but shrink the set \(\mathcal U\) of future runs over which savings accumulate. The model gives two design goals: \emph{maximize reuse} across workloads that benefit from an edit, and \emph{minimize the time needed to verify} that benefit. Extending reuse pays off when the additional runs save more time than they cost to verify.

For a fixed model and hardware, workloads with different token values can execute the same operations on similarly shaped tensors. Systems-only search assumes that an implementation speedup measured on one such workload transfers to the others~\cite{researchboost,infrarsi}. It also assumes that the candidate and reference reach the quality target in the same number of updates (\cref{eq:system-objective}). Multiplying that count by the mean duration of a few representative updates estimates total training time. Agreement with the reference on those same updates provides a low-cost proxy for full-run fidelity.

Allowing the learning rule to change introduces challenges for both reuse and verification in \cref{eq:net-savings}. For reuse, the same edit can reduce the number of updates needed to reach the target on one dataset and increase it on another, even when tensor shapes match. For verification, a candidate can diverge from the reference on its first update yet reach the target sooner, or agree for \(m\) updates and later miss the target. Confirming a speedup on a new workload may thus require a full run, whose cost must be recovered through reuse.

\Needspace{4\baselineskip}
\section{A-Train}
\label{sec:system}

Given a reference trainer and a workload configuration, \atrain{} searches for optimizations whose cumulative savings over repeated runs can offset the cost of search (\cref{eq:net-savings}). It returns an optimized trainer for reuse across unseen workload dimensions. We begin with an overview of the agent loop that guides the search.

\subsection{A Scalable Loop for Full-Stack Optimization}

\begin{figure}
\centering
\includegraphics[width=\linewidth]{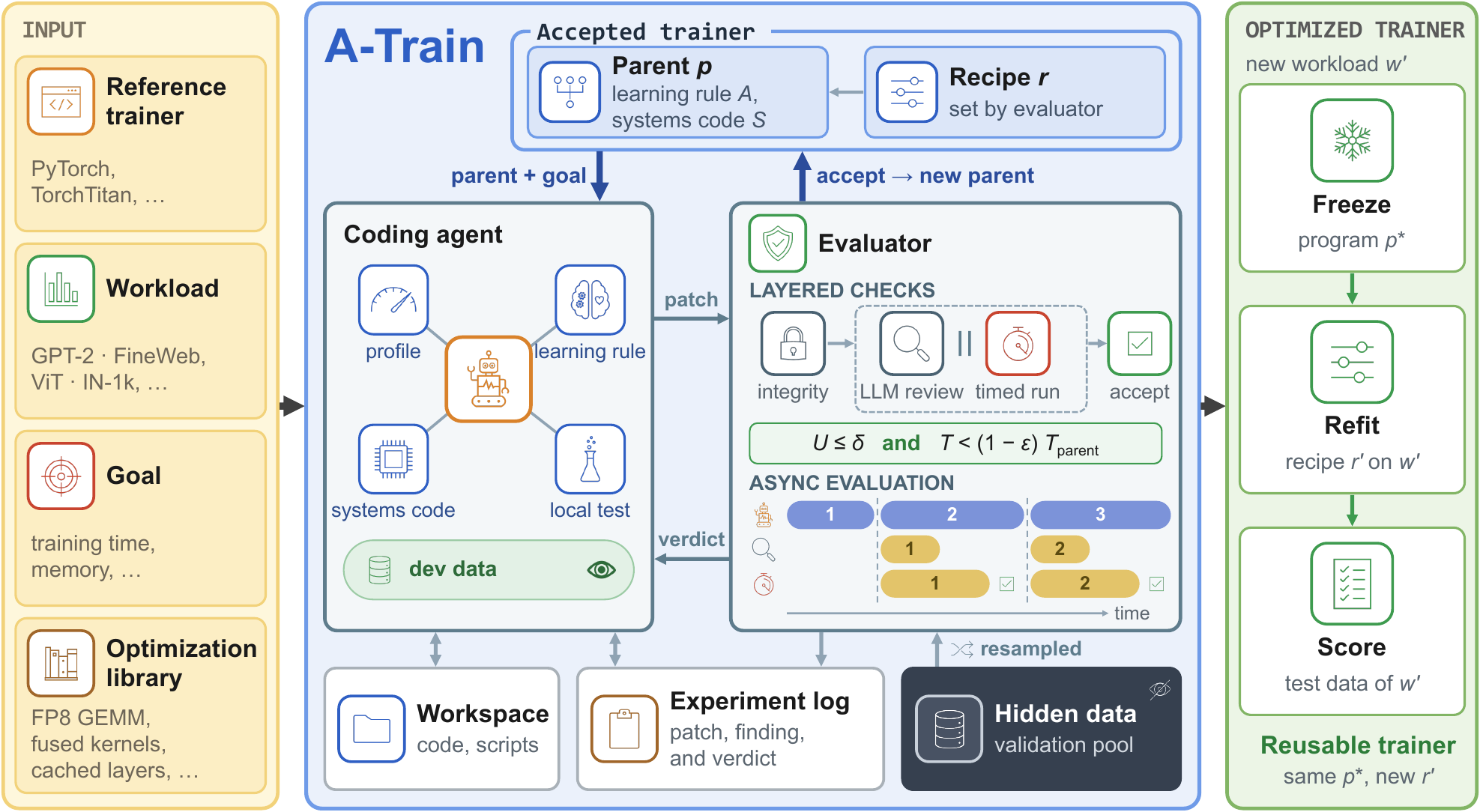}
\caption{\textbf{One round of trainer co design.} The agent develops an edit locally. The evaluator runs it on hidden data and returns a coarse verdict. After search, the program is frozen and its recipe can be fitted to a new workload.}
\label{fig:method}
\end{figure}

As coding agents continue to improve with scaling law, we expect them to make better decisions about what to optimize and how to optimize it.
Prior work suggests that encoding human heuristics in the loop can restrict an agent's action space and limit achievable speedups~\cite{togethercompiler}. 
To let the optimizer scale with its agent's capability, we design a minimally sufficient loop that supplies only the necessary artifacts and leaves optimization decisions entirely to the agent (\cref{fig:method}).

Each round starts with the accepted trainer (the \emph{parent}). Given an optimization goal, the agent decides what to investigate and how to proceed. For example, it might profile the parent, modify the learning rule or systems code, or test a change on development data. It can submit a patch, carry unfinished work into the next round, or record a finding. A separate evaluator checks each submitted patch, provides feedback, and decides whether to accept it. If accepted, the updated trainer becomes the parent for the next round.

\noindent\textbf{Agents choose their own workflow.}
Prior optimization loops assign planning, implementation, and performance diagnosis to distinct roles~\citep{vibeserve,infrarsi}. Across repeated runs, we found that stronger coding agents could coordinate these activities on their own. This observation motivates a simpler design for \atrain{}: a single agent receives an optimization goal and feedback and chooses its own next step. With this guidance alone, agents initiated bottleneck investigations, devised tests for their assumptions, and revised their implementations without step-specific prompts or a prescribed workflow. The agent also uses local tests to decide which edits warrant a full training run on the hidden workload.

\noindent\textbf{What must survive a round?}
To sustain learning across optimization rounds, we preserve only the context an agent needs to connect prior attempts with feedback from the environment and choose its next step. The \emph{workspace} retains scripts and unfinished code, while the \emph{experiment log} links each submitted patch to the agent's findings and the evaluator's verdict. An agent used both to recover a rejected decoder rewrite, correct its sample order, and resubmit it successfully. Losing the patch would have cost agent time and tokens to reconstruct it. Losing the verdict could have led the agent to retry the failed version. We argue that even stronger agents will need both forms of state.

\noindent\textbf{What can carry over to a new search?}
While specializing to a workload reveals new optimization opportunities, we find that some recur across workloads of shared structures. We use an \emph{optimization library} of verified patches from earlier searches to warm start a new search. Each entry records a reference patch, how it works, its measured effect, and the conditions under which it helped or failed. The loop ranks entries by their fit to the new trainer's profile. The agent chooses which ideas to adapt, and the evaluator tests each resulting patch on the new workload.

\Needspace{4\baselineskip}
\subsection{A Regularized Search for Reusable Trainers}

A useful trainer should generalize to workloads that vary along an axis absent from the configurations used for optimization. To encourage this, we regularize program search by limiting both the training decisions the agent can encode and the information it can observe when choosing optimizations.

\noindent\textbf{Structural regularization.}
The agent writes a program $p=(\algorithmSymbol,\systemSymbol)$ that takes a training recipe $r$ as input. We define $r$ as the designated set of workload-dependent hyperparameters that the evaluator controls. Without this separation, we found that agents spent much of the search tuning these hyperparameters for easy gains. This invites reward hacking because recipe tuning can improve the score on the search workloads without making $p$ more reusable. To keep the search focused on program changes, the evaluator supplies $r$ and verifies that $p$ honors its values during execution. The evaluator then refits $r$ on a new workload while reusing $p$.

\noindent\textbf{Feedback regularization.}
Structural regularization only targets recipe overfitting. To mitigate program overfitting, we create an information gap between development and validation. The agent can inspect detailed diagnostics on development workload. On validation workload, the evaluator hides examples and traces, reporting only unmet requirements and coarse differences in timing and loss. For each fixed workload $w$, the evaluator also resamples validation data from a hidden pool, allowing the empirical distribution to vary across evaluations and reducing the risk that successive edits overfit a single distribution.

\Needspace{4\baselineskip}
\subsection{A Faithful and Efficient Evaluator}

Our evaluator measures how much training time each optimization saves while meeting the required quality target. To keep this measurement faithful, we use \textbf{layered} checks to verify that each edit satisfies the task requirements and reduces training time relative to the currently accepted trainer. To keep evaluation practical across many search rounds, we run these checks \textbf{asynchronously} so agents can continue searching for optimizations while verification is underway.

\noindent\textbf{Layered checks for faithful measurement.}
We evaluate each candidate in three layers to establish that the measurement procedure is intact, the optimization follows the training protocol, and the resulting trainer is more efficient while passing the quality gate. 
First, the evaluator checks the integrity of evaluator owned files to detect any attempt to modify the measurement procedure. 
Second, an LLM reviewer inspects the patch for protocol violations that require interpreting the candidate's code, such as truncating training sequences or switching models. 
Third, an official run measures the candidate's training time and model quality on the hidden workload using its parent's recipe. 

Using the same prescribed reference for every optimization prevents cumulative quality loss (\cref{app:loss}). We accept a valid trainer when
\begin{equation}
  U(p,r;\mathrm{REF}) \leq \delta
  \qquad\text{and}\qquad
  T(p,r) < (1-\epsilon)\,T_{\mathrm{parent}},
  \label{eq:accept}
\end{equation}

where $U$ bounds the quality gap between the candidate and the reference, $\delta$ is the workload's quality tolerance, and $\epsilon$ is the required fractional reduction in training time relative to the latest accepted parent.

\noindent\textbf{Asynchronous evaluation for efficient search.}
Waiting for each evaluation to finish leaves the agent idle. We reduce this delay by overlapping work within and across rounds. Within a round, code review and the official training run proceed in parallel, and recipe trials can start speculatively. If a valid trainer misses an acceptance condition at the inherited recipe, the evaluator continues the refit only when the run leaves quality equivalence to the parent unresolved. A fresh timed run tests the selected recipe. 

Across rounds, the agent can develop its next patch while a refit is pending. Since the refit can change both the code the patch modifies and the time it must beat, the evaluator decides on optimizations in submission order. If the earlier optimization advances, the evaluator applies the follow up patch to the new parent and returns any conflicts to the agent for repair. It then tests the patch at the new parent's recipe and measures its speedup against that parent.

\definecolor{specaxis}{HTML}{006970}
\definecolor{reuseaxis}{HTML}{934315}
\newcommand{\specval}[1]{\textcolor{specaxis}{#1}}
\newcommand{\reusepath}[3]{\textcolor{reuseaxis}{\begin{tabular}[c]{@{}l@{}}#1\,$\to$\,#2\\$\to$\,#3\end{tabular}}}

\begin{table}[t]
\centering
\small
\setlength{\tabcolsep}{3pt}
\caption{\textbf{Specialization and reuse across the six searches.}
\textcolor{specaxis}{Single values} are fixed by the workload
configuration \(Z\). \textcolor{reuseaxis}{Arrows} trace each varied dimension
from development through evaluator feedback to the frozen trainer's main test.}
\label{tab:settings}
\begin{tabular*}{\linewidth}{@{\extracolsep{\fill}}clll@{}}
\toprule
Scenario & Model & Hardware & Data \\
\midrule
A & \specval{Qwen3-1.7B}       & \specval{HBM3 GPU}
  & \multirow{3}{*}{\reusepath{Chat and summaries}{Evol-CodeAlpaca}{Magicoder-OSS-Instruct}} \\
B & \specval{Qwen3-1.7B}       & \specval{Consumer GPU} & \\
C & \specval{Granite-4.0-H-1B} & \specval{HBM3 GPU}     & \\
\cmidrule(l){4-4}
D & \specval{Whisper-base} & \specval{Xeon CPU}
  & \reusepath{LibriSpeech}{VoxPopuli}{held-out VoxPopuli speakers} \\
\midrule
E & \specval{Qwen3-1.7B} & \reusepath{HBM3 GPU}{HBM2e GPU}{consumer GPU}
  & \multirow{2}{*}[-.5\normalbaselineskip]{\reusepath{Chat and summaries}{Evol-CodeAlpaca}{Magicoder-OSS-Instruct}} \\
F & \reusepath{Qwen3-1.7B}{Llama-3.2-1B}{Granite-3.1 and Granite-4.0-H} & \specval{HBM3 GPU} & \\
\bottomrule
\end{tabular*}
\end{table}

\section{Evaluation}
\label{sec:evaluation}

We evaluate \atrain{} as a foundation for agent-driven optimization of the training stack by measuring how much training time an agent can save with feedback from a single workload. To probe the limits of workload specialization, we test scenarios with diverse devices, models, and datasets. 

\subsection{Experimental Setup}
\label{sec:setup}

\noindent\textbf{Workloads.} \Cref{tab:settings} shows which workload dimensions each search fixes for specialization and which it varies to test reuse. For Qwen3~\citep{qwen3} and Granite~\citep{granite4nano}, we use chat and document summaries for development, Evol-CodeAlpaca~\citep{evolcodealpaca} code instructions for evaluator feedback, and deduplicated Magicoder-OSS-Instruct~\citep{magicoderdata} examples for the main test, alongside three additional held out tests covering a shifted chat and paper mix, a small data split, and long documents. For Whisper~\citep{whisper}, we use LibriSpeech audiobooks~\citep{librispeech} for development and VoxPopuli parliamentary speech~\citep{voxpopuli} for feedback, testing transfer on held out VoxPopuli speakers plus LibriSpeech Other, LibriSpeech 360\,h, and a small data split. The language models run on two 80\,GB datacenter GPUs (HBM3 and HBM2e) and a 32\,GB GDDR7 consumer GPU (\cref{tab:gpu-specs}), while Whisper runs on a Xeon CPU.

\noindent\textbf{Baselines and implementations.} We compare against PyTorch reference trainers built with Hugging Face Transformers (TRL \texttt{SFTTrainer} for LLMs and \texttt{Seq2SeqTrainer} for Whisper) and, in Scenario~A, torchtitan, each tuned independently on every workload with the same evaluator recipe search over learning rate, schedule shape, decay fraction, learning rate floor, and step budget (\cref{app:refit}). We also compare \atrain{} with systems only search on Scenarios~A and~B. We found no suitable production baseline for Qwen3 on the consumer GPU, Whisper on the Xeon CPU, or Granite on the HBM3 GPU. All searches use Claude Code and a 20 agent hour budget (\cref{app:refit}); \cref{sec:analysis} reports search costs.

\subsection{Training Performance and Co Design Gains}
\label{sec:results}
\label{sec:reuse}

\begin{figure}
\centering
\includegraphics[width=\linewidth]{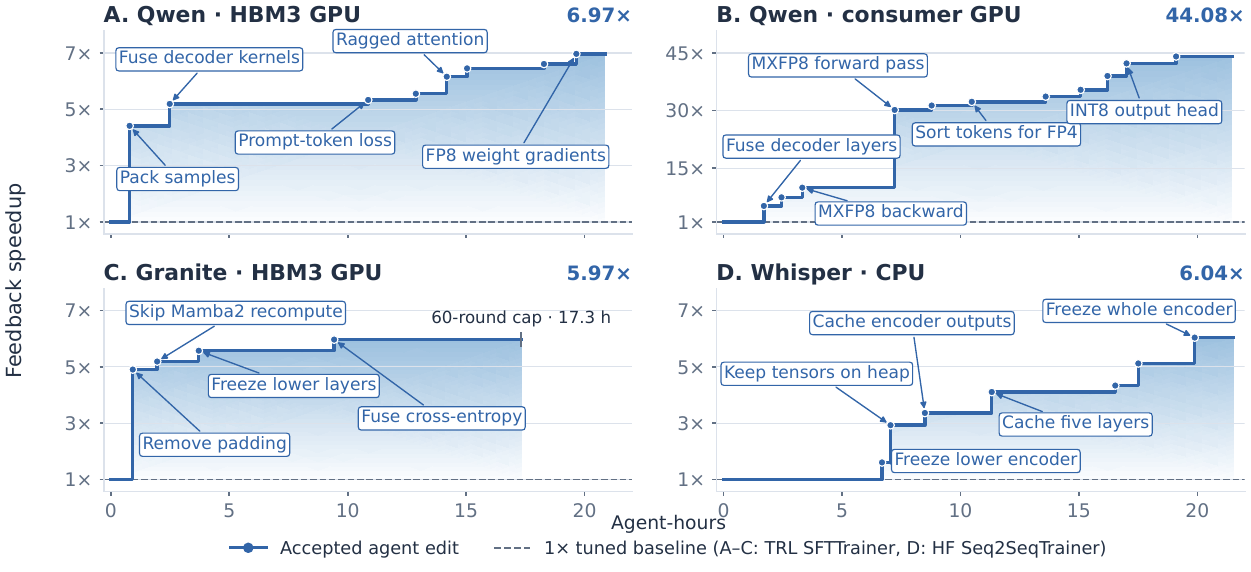}
\caption{\textbf{Search trajectories (Scenarios~A--D).} Dots show accepted edits vs.\ tuned baselines.}
\label{fig:data-search}
\end{figure}

\begin{figure}
\centering
\includegraphics[width=\linewidth]{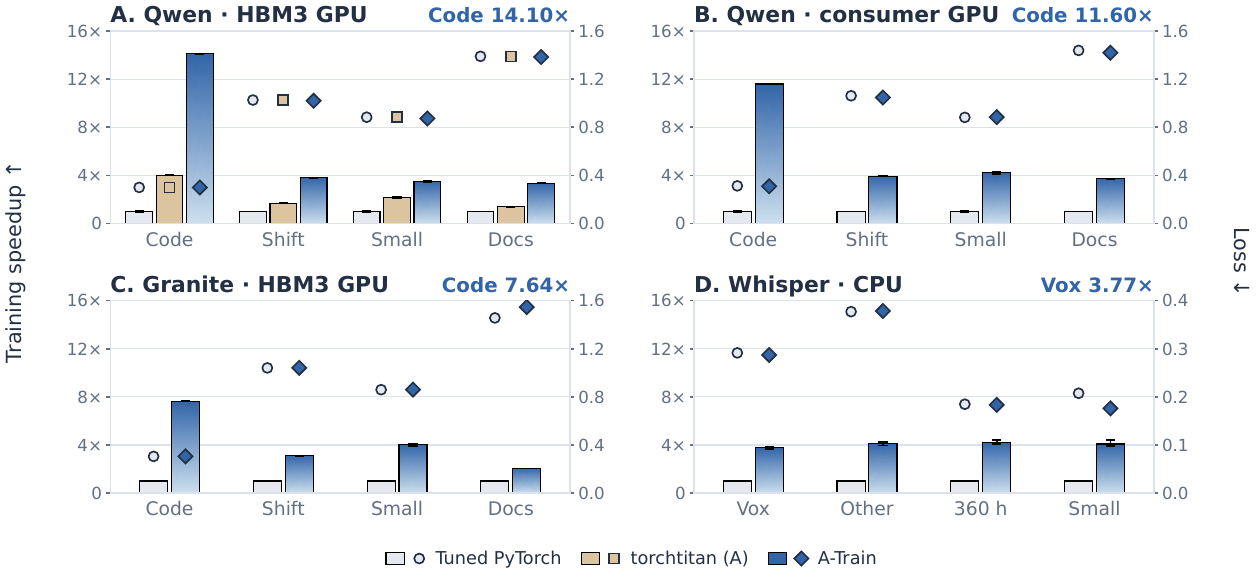}
\caption{\textbf{Held out speedup and test loss (Scenarios~A--D).} Bars show speedup over tuned PyTorch; markers show mean test loss.}
\label{fig:transfer}
\end{figure}

We first evaluate the four scenarios that specialize a trainer to a fixed model and device while transferring across datasets. \Cref{fig:data-search} traces accepted edits during search, and \Cref{fig:transfer} reports held-out speedup and test loss after re-tuning each frozen trainer on fresh data.

\noindent\textbf{Scenario~A: Dense LLM SFT on an HBM3 GPU.}
The agent first strips away system overhead by packing whole samples into microbatches and using fused decoder kernels, reaching $5.19\times$ on feedback. It then makes a co-design edit, adding an auxiliary prompt-token loss that slightly raises per-step cost but reaches the loss target in fewer steps (\cref{sec:discoveries}). Layering ragged cuDNN attention, 16-bit AdamW moments, and scale-folding FP8 weight gradients (\cref{sec:novel}) brings feedback speedup to $6.97\times$. On the held-out code test, the frozen trainer finishes in 229\,s versus 3,224\,s for the tuned reference, achieving a $14.10\times$ speedup, while tuned torchtitan finishes in 803\,s for a $4.01\times$ speedup ($3.51\times$ slower than \atrain{}). Held out speedup exceeds feedback speedup ($14.10\times$ versus $6.97\times$) because refitting cuts training from 1,854 to 1,424 steps while packing removes more padding on Magicoder. It also meets the quality target on held-out datasets.

\noindent\textbf{Scenario~B: Low-precision co-design on a consumer GPU.}
On the 32\,GB GDDR7 GPU, memory-lean fused decoder layers and packed microbatches eliminate full gradient checkpointing to reach $7.41\times$ on feedback, freeing the batch sizes needed to saturate low-precision tensor cores. MXFP8 forward/backward GEMMs with stochastically rounded BF16 AdamW jump to $30.16\times$ once refitted, and NVFP4 gradient GEMMs stabilized by permutation sorted quantization (\cref{sec:novel}) alongside 8-bit attention and sparse LM-head kernels (\cref{app:novel}) reach $44.08\times$. On held-out code, the frozen trainer finishes in 204\,s for an $11.60\times$ speedup, sustaining $3.73$--$4.23\times$ across the other three workloads while meeting every loss target. 

\noindent\textbf{Scenario~C: Hybrid Mamba-2/attention model on an HBM3 GPU.}
Granite interleaves Mamba-2 state-space layers with periodic self-attention, making state-space scans and MLPs dominate step time. Padding-free packing across both layer types yields $4.91\times$, while removing Mamba-2 backward recomputation, fusing cross-entropy, and restricting the top Mamba-2 layer to loss-bearing tokens further reduce step time. Freezing the bottom 4 of 40 layers reaches $5.97\times$ on feedback. On held-out workloads, the frozen trainer achieves speedups of $7.64\times$ on code, $4.03\times$ on small data, $3.12\times$ under distribution shift, and $2.07\times$ on long documents.

\noindent\textbf{Scenario~D: Multi-epoch ASR on a Xeon CPU.}
Retaining large tensors on the glibc heap eliminates per-allocation \texttt{mmap}/\texttt{munmap} syscalls, cutting runtime by 45\,\% and bringing feedback speedup to $2.93\times$. Incrementally freezing the acoustic encoder then unlocks cross-epoch activation caching: the agent caches five layers, splits top-encoder attention from its feed-forward block, and ultimately freezes and caches the full encoder while keeping decoder cross-attention key/value projections trainable to reach $6.04\times$ on feedback. On held-out VoxPopuli speakers, the frozen trainer runs $3.77\times$ faster than the tuned reference and sustains $4.08$--$4.23\times$ across the other three speech workloads while meeting every loss target. On small-data ASR, encoder freezing also regularizes training, lowering word error rate from 6.31\,\% to 5.39\,\%.

\begin{figure}[t]
\centering
\includegraphics[width=\linewidth]{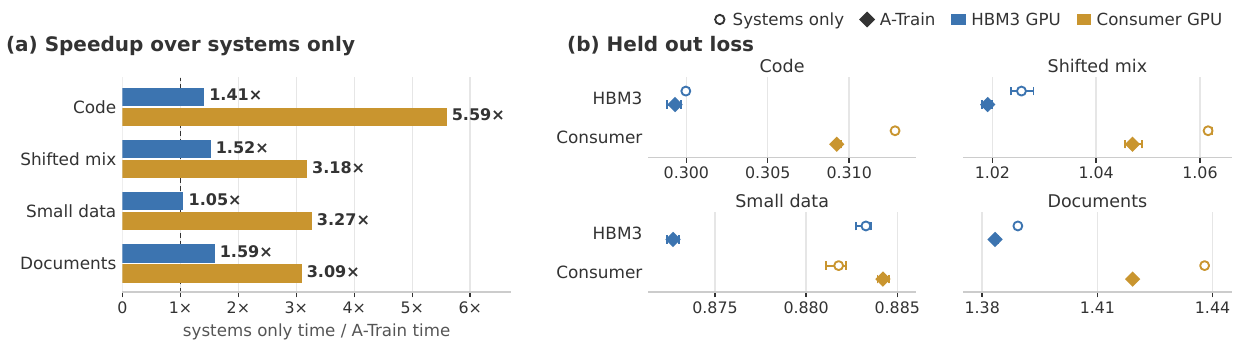}
\caption{\textbf{Co design versus systems only search on Qwen3-1.7B (Scenarios~A and~B).} \textbf{(a)}~Held out speedup of \atrain{} over systems only search. \textbf{(b)}~Held out loss.}
\label{fig:codesign}
\end{figure}

\noindent\textbf{Co design versus systems only search.}
Expanding the search to modify the learning algorithm consistently outperforms exact systems only rewrites across all four held out Qwen3 workloads in Scenarios~A and~B (\cref{fig:codesign}). On the HBM3 GPU, co design delivers $1.05$--$1.59\times$ faster training at lower mean loss on every workload. The advantage widens substantially on the memory constrained consumer GPU ($3.09$--$5.59\times$), where algorithmic changes unlock low precision tensor cores while still meeting every loss target.

\subsection{Adaptation across Hardware and Models}
\label{sec:adaptation}

\begin{figure}
\centering
\includegraphics[width=\linewidth]{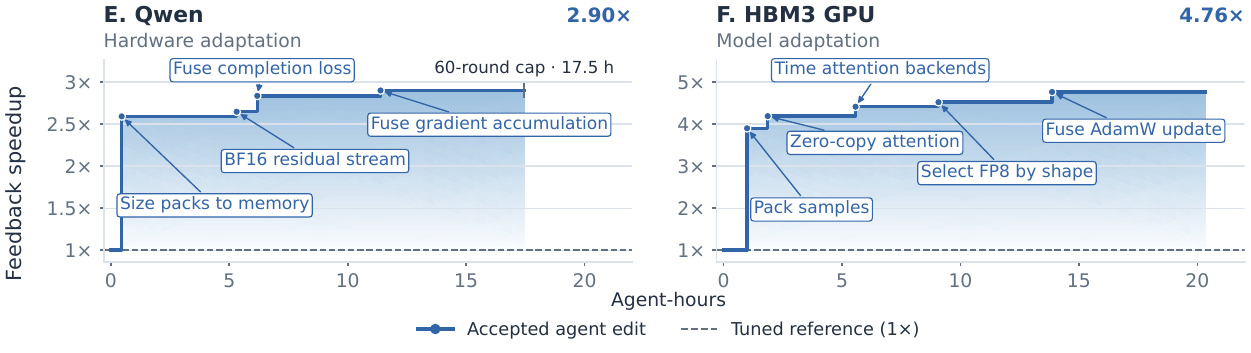}
\caption{\textbf{Adaptation search trajectories (Scenarios~E--F).}}
\label{fig:adaptation-search}
\end{figure}

\begin{figure}
\centering
\includegraphics[width=\linewidth]{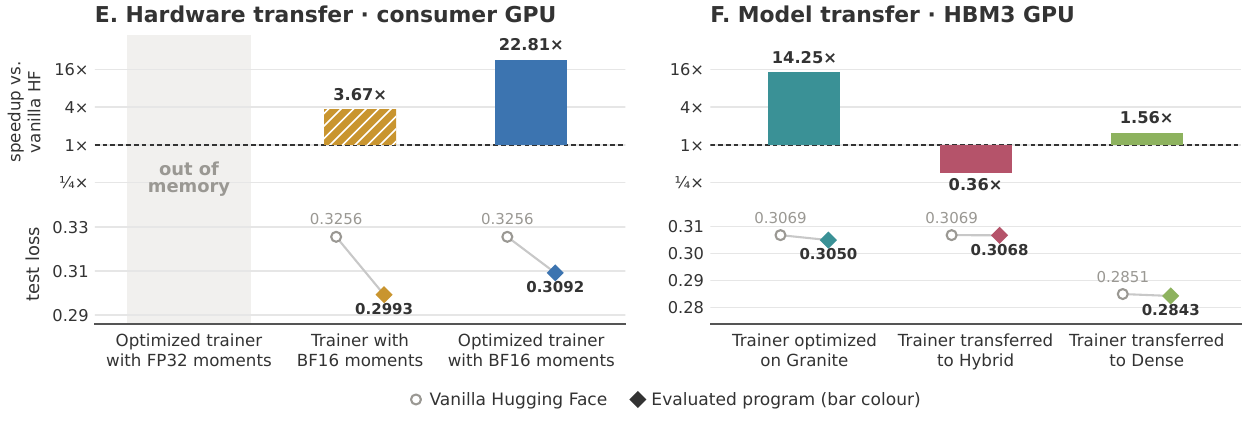}
\caption{\textbf{Hardware and model transfer vs.\ vanilla Hugging Face (Scenarios~E--F).}}
\label{fig:adaptation-tests}
\end{figure}

We next test generalization to unseen GPUs and model architectures (\cref{fig:adaptation-search,fig:adaptation-tests}). Both searches develop on Qwen3-1.7B on an HBM3 GPU, but receive feedback on a different target: HBM2e in Scenario~E and Llama-3.2-1B in Scenario~F. This steers search from hardcoded kernels to runtime self calibration (\cref{sec:discoveries}).

\noindent\textbf{Scenario~E: Hardware adaptation across GPU generations (HBM3 $\to$ HBM2e $\to$ consumer GPU).}
Because HBM2e lacks FP8 tensor cores and has lower bandwidth (\cref{tab:gpu-specs}), the agent focuses on hardware-portable optimizations: runtime memory probes dynamically size packed-token budgets and checkpointing depth to reach $2.59\times$ on HBM2e, while BF16 residual streams, fused completion loss, and GEMM-epilogue FP32 gradient accumulation bring feedback speedup to $2.90\times$. To probe the limits of reuse, we deliberately test the frozen trainer on a 32\,GB consumer GPU, a memory constraint absent during search since both the HBM3 and HBM2e GPUs have 80\,GB. As a result, FP32 weights, gradients, and AdamW moments alone occupy $\approx$30.5\,GB and cause an OOM, showing that the trainer cannot generalize to constraints it never encountered. Manually moving optimizer moments to BF16 lets the remaining optimizations transfer, finishing in 1,267\,s versus 2,459\,s for the tuned reference ($1.94\times$, or $3.67\times$ over vanilla) at lower test loss.

\noindent\textbf{Scenario~F: Model adaptation across dense and hybrid architectures (Qwen3 $\to$ Llama-3.2 $\to$ Granite).}
Switching feedback to Llama-3.2-1B breaks Qwen-specific rewrites, so the agent embeds startup microbenchmarks and correctness guards: it times per-sample cuDNN against varlen FlashAttention to set a sequence-length switchover, gates FP8 input-gradient GEMMs on per-shape speedups, and checks that packing preserves hidden states, reaching $4.76\times$ on Llama feedback. On unseen dense Granite-3.1, these checks succeed and yield $1.56\times$ over vanilla, whereas Scenario~A's Qwen trainer crashes before step~1 on both Granite models. On hybrid Granite-4.0-H, however, the equivalence check catches Mamba-2's cross-example state leakage and disables packing, avoiding a crash and preserving loss parity at the cost of a steep slowdown to $0.36\times$. Because validation used only dense Transformers, this result shows that robustness can transfer to unseen architectures, whereas speedup remains bounded by the architectural diversity covered during validation.
\subsection{Novel Methods Discovered by Agents}
\label{sec:novel}

Across our searches, agents discovered specialized techniques spanning OS allocators, custom kernels, low precision arithmetic, activation caching, learning objectives, and runtime guards (\cref{tab:discoveries}).
We highlight three examples from distinct workloads here, with two additional mathematical methods from Scenario~B, exact algebraic identities for consistent $\Delta$ 8 bit FlashAttention and gradient asymmetry in active vocabulary head updates, detailed in \cref{app:novel}.

\begin{table}[t]
\centering
\footnotesize
\setlength{\tabcolsep}{4pt}
\renewcommand{\arraystretch}{1.08}
\caption{\textbf{Summary of full stack discoveries across Scenarios~A--F.} Each search combines optimizations across distinct layers of the training stack.}
\label{tab:discoveries}
\begin{tabularx}{\linewidth}{@{}l l Y@{}}
\toprule
Scenario & Stack layers & Signature discoveries \\
\midrule
A (Qwen3, HBM3) & Kernels, precision, objective & Auxiliary prompt token loss, ragged cuDNN attention, 16 bit AdamW moments, scale folding FP8 weight gradients \\
B (Qwen3, 32\,GB) & Memory, kernels, precision & Permutation sorted NVFP4 gradients, MXFP8 GEMMs with stochastic BF16 AdamW, consistent $\Delta$ 8 bit FlashAttention, active vocabulary head updates \\
C (Granite, HBM3) & Kernels, architecture & Padding free Mamba 2 and attention packing, top layer loss row restriction, bottom layer freezing \\
D (Whisper, CPU) & OS allocator, caching, objective & glibc heap retention cutting mmap page faults by 45\,\%, frozen acoustic encoder with cross epoch activation caching \\
E (Hardware axis) & Runtime probes, memory & Dynamic startup memory probes for packing and checkpointing depth, GEMM epilogue FP32 gradient accumulation \\
F (Model axis) & Runtime guards, compilation & Zero weight compile twin, startup attention backend switchover, shape gated FP8, packing equivalence guard \\
\bottomrule
\end{tabularx}
\end{table}

\noindent\textbf{Permutation sorted FP4 weight gradients (Scenario~B).}
NVFP4 shares one scale across every 16 tokens, so one large gradient token pushes the other 15 into rounding noise, causing 25\,\% local gradient error.
Because weight gradients sum over tokens in any order, the agent sorts tokens per layer by gradient magnitude $\lVert dY_t \rVert_\infty$ before quantizing both operands, $dW = \sum_t dY_{\pi(t)}^\top X_{\pi(t)}$, cutting error to 9.3\,\% without altering the exact product.

\noindent\textbf{Scale folding FP8 weight gradients (Scenario~A).}
Running weight gradients $dW = dY^\top X$ in FP8 normally requires quantizing both operands anew along the token axis, eroding tensor core gains.
Instead, the agent reuses an existing per token FP8 gradient copy $\hat{dY}$ (with token scales $s_t$) untouched and folds $s_t$ into the paired activation before channel quantization: $dW = \sum_t \hat{dY}_t^\top (s_t X_t)$, cutting feedback runtime by 5.1\,\%.

\noindent\textbf{Zero weight compile twin (Scenario~F).}
When the target architecture is unknown in advance, compiling training graphs at step 0 adds heavy startup delay.
At launch, the agent forks a shadow process on a zero weight copy built from the model configuration alone to precompile the graph into a private cache while the main process loads real weights; after verifying matching setup signatures, step 0 reuses the cached kernels, cutting feedback runtime by 5.1\,\%.

\subsection{Behavior Analysis}
\label{sec:discoveries}

\noindent\textbf{A single agent optimizes across the entire stack, uniting domains that normally require separate experts.}
Rather than stopping at a specific stack, the agents follow unexplained latency or loss wherever it leads, from OS kernels and hardware limits to software bugs and training objectives.
In Scenario~D, a Whisper encoder layer took far longer than the sum of its operators; the agent traced the gap to repeated memory allocation page faults and tuned the system allocator, cutting feedback runtime by 45\,\%.
On the consumer GPU in Scenario~B, overlapping optimizer updates with forward passes looked fast in isolation yet failed in training; the agent found that concurrent memory traffic hit the card's power limit and throttled matrix multiplies.
In Scenario~F, reordering two attention calls exposed severe gradient corruption inside the accepted trainer, which the agent traced to a caching bug in cuDNN and fixed.
Moving upward into the learning objective, the agent in Scenario~A added a lightly weighted loss on prompt tokens that slowed every step, betting that it would finish training sooner.

\noindent\textbf{The agent behaves like a skeptical scientist, falsifying its own evidence and revising past choices.}
Rather than trusting noisy benchmarks or locking in early wins, the agents repeatedly test and overturn their own findings.
In Scenario~B, the agent questioned gradient error metrics in both directions: it kept an FP4 backward pass whose large batch error barely affected training loss, yet aborted a sparse head speedup with tiny aggregate error after finding that subtle bias degraded local validation loss.
In Scenario~E, after many stalled rounds had dismissed gradient fusion on assumed Python overhead, a direct control experiment showed the overhead was hidden by GPU compute and unlocked a speedup.
In Scenario~D, the agent even reversed an earlier freeze of decoder attention projections once freezing the full encoder required that capacity to adapt.

\noindent\textbf{The agent engineers for unseen workloads, synthesizing diverse benchmarks and trainers that adapt at runtime.}
Because target data, models, and devices can differ from development, the agents build varied local test suites rather than tuning only to the nominal dataset.
In Scenario~F, after attention edits failed to transfer to feedback, the agent built shorter subsets of its training data, found that different attention backends won at short and long lengths, and added a startup benchmark to pick the switchover point.
In Scenario~B, the agent created shorter and fully scored variants of its development data to stress memory peaks and output layer costs, reproducing a hidden memory failure locally.
To avoid overfitting to one model or GPU, the Scenario~F agent also tested ten randomly initialized proxy architectures, catching a severe regression invisible on Qwen3, and shipped a packing equivalence check that later caught state leakage on unseen hybrid Granite; similarly, the Scenario~E agent added startup memory probes to size packing and checkpointing on unseen GPUs.
In conclusion, with a regularized search, the agent tries to maximize the reusability of the optimized trainers with adaptive startup checks and correctness guards (\cref{fig:transfer,fig:adaptation-tests}).

\subsection{Library Warm Start and Search Cost}
\label{sec:analysis}

\begin{figure}[t]
\centering
\includegraphics[width=\linewidth]{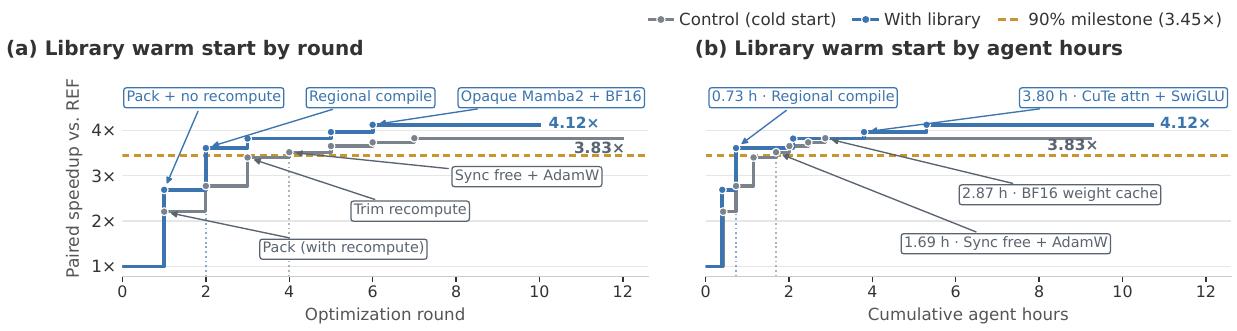}
\caption{\textbf{Optimization library warm start on Granite-4.0-H-1B.} Granite search warm started with a 15 entry Qwen3-1.7B optimization library versus cold start, tracked by \textbf{(a)}~round and \textbf{(b)}~agent hours.}
\label{fig:library}
\end{figure}

\noindent\textbf{Warm starting with the optimization library.}
Warm-starting a systems-only Granite-4.0-H-1B search with a 15-entry library extracted solely from earlier Qwen3-1.7B runs accelerates early discovery and raises final performance, as shown in \cref{fig:library}. Across two seeds, warm-starting halves the rounds and more than halves the agent-hours needed to reach 90\,\% of the cold-start control's speedup, and yields higher final full-run speedups of $3.56\times$ and $3.72\times$ versus $3.39\times$ while passing all three-seed quality checks. Rather than copying patches verbatim, every accepted edit adapts or combines library ideas for the hybrid architecture while the evaluator filters out non-transferring attempts. Although fast cold start convergence relative to single seed variance precludes a formal significance claim, both seeds find stronger trainers sooner.

\begin{figure}
\centering
\includegraphics[width=\linewidth]{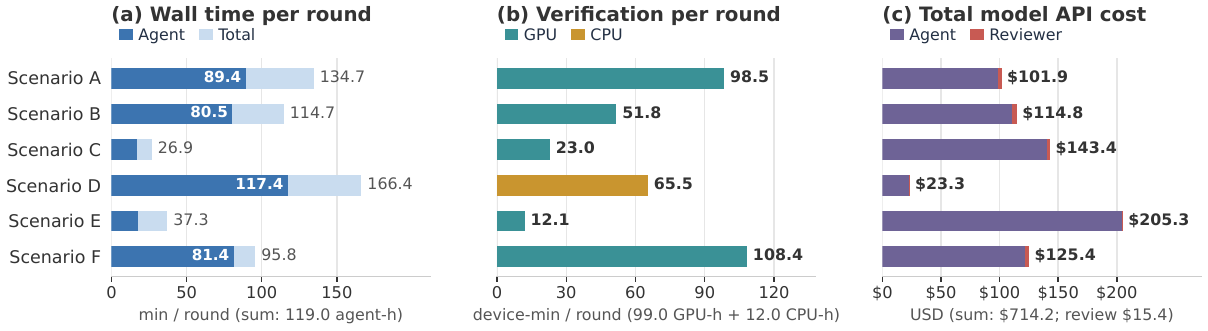}
\caption{\textbf{Search cost across Scenarios~A--F.} \textbf{(a)}~Mean active agent vs.\ total wall time per round. \textbf{(b)}~Mean parallel verification compute per round. \textbf{(c)}~Total model API spend.}
\label{fig:cost}
\end{figure}

\noindent\textbf{Search cost and amortization.}
\Cref{fig:cost} shows that total wall time per round remains close to active agent time even when parallel verification consumes comparable or greater device-time, confirming that asynchronous evaluation hides most verification latency. Scenarios~C and~E average shorter rounds because both hit the 60-round cap with many brief no-change turns that skipped evaluation. Across all six searches, Layer~2 patch review adds only about 2\,\% of reported model API spend. Under \cref{eq:net-savings}, upfront verification compute amortizes rapidly on longer workloads. In Scenario~A, 23.0 HBM3 verification GPU hours (\$102 API spend) break even in 15 long document runs or 28 code runs, saving 60 to 134 GPU hours over 100 runs; in Scenario~B, 13.8 consumer GPU hours (\$115 API spend) break even in 6 document runs or 23 code runs, saving 46 to 206 GPU hours over 100 runs; and in Scenario~D, 12.0 CPU hours (\$23 API spend) break even in 34 VoxPopuli runs. These comparisons illustrate device time amortization rather than full economic break even, as reported model totals exclude hardware charges, missing turn level usage logs, and post search audits.

\section{Related Work}
\label{sec:related}

\paragraph{Agentic optimization.}
LLM and coding agents increasingly automate optimization across diverse targets, including prompts~\citep{opro,textgrad,gepa}, agent harnesses and workflows~\citep{adas,vesper,xia2026rrsi}, accelerator kernels~\citep{cudaforge,drkernel,accelopt,autokernel}, inference and serving engines~\citep{vibeserve,metainfer,seis,togethercompiler}, and scientific algorithms~\citep{alphaevolve}. Across these domains, verification is comparatively cheap: prompts and harnesses are scored on fixed validation sets, while kernel and serving optimizations preserve the underlying computation so output equivalence and short microbenchmarks suffice. Optimizing ML training is harder because altering the learning algorithm changes convergence over full runs, coupling step speed with sample efficiency and requiring costly full validation.

\paragraph{Agentic training.}
Agentic training optimizers follow two main lines. Systems agents, including ASAP~\citep{asap}, ATTUNE~\citep{attune}, ResearchBoost~\citep{researchboost}, and InfraRSI~\citep{infrarsi}, tune parallelism, configurations, kernels, or trainer infrastructure for step throughput while fixing the learning algorithm, so short timing and numerical or loss checks suffice. Research agents, including autoresearch~\citep{autoresearch,bilevel_autoresearch}, Athanor~\citep{athanor}, and EvoTrainer~\citep{evotrainer}, alongside ML engineering, finetuning, and speedrunning benchmarks~\citep{mlebench,rebench,speedrun_bench,posttrainbench}, modify training scripts or harnesses, typically maximizing quality under a fixed compute budget or targeting a single workload. To our knowledge, \atrain{} is the first work to jointly optimize the full training stack across both the learning algorithm and systems implementation to minimize total time to a quality target~\citep{algoperf}.

\par\addvspace{0.65\baselineskip}\noindent\begin{minipage}{\linewidth}
\section{Conclusion}

We presented \atrain{}, demonstrating that a single coding agent can autonomously co-design a trainer's learning algorithm and systems implementation into specialized, reusable training code.
Across six language and speech searches spanning datacenter GPUs, a consumer GPU, and a CPU, \atrain{} delivers \mbox{$3.77$--$14.10\times$} held-out speedups over independently tuned PyTorch references, outperforms tuned torchtitan and systems-only search, and discovers novel algorithms alongside runtime self-calibration guards.
By pairing a minimal agent loop with regularized search and evaluator-controlled recipes, \atrain{} achieves these gains at modest API spend, with verification compute amortizing over moderate reuse.
Looking ahead, as coding agents improve and verification scales to distributed training, autonomous full-stack co-design offers a scalable lever for lasting training efficiency.
\end{minipage}\par

\FloatBarrier
{\footnotesize\setlength{\bibsep}{0.5pt}
\bibliographystyle{plainnat-compact}
\bibliography{main}}

\clearpage
\beginappendix

\section{Feedback Protocol and Hardware}
\label{app:protocol}

\subsection{Feedback Loss}
\label{app:loss}

For feedback example $i$, let $d_i$ be the candidate's summed loss minus the mean summed loss over reference seeds. With $n$ examples and $W$ scored tokens, the one-sided 95\% upper bound on excess loss per token is
\[
  U_w = \frac{\sum_i d_i + t_{0.95,n-1}\sqrt{n}\,\mathrm{sd}(d_i)}{W}.
\]
The evaluator requires $U_w\leq\delta_w$. 
At the inherited recipe, losses are equivalent if the paired 90\% interval against the parent lies strictly inside $(-\delta_w/4,\delta_w/4)$.

\subsection{Recipe Refitting and Agent Setup}
\label{app:refit}

All searches run with Claude Code as the coding agent under a budget of about 20 agent hours, with local access only to development data while official feedback and final test splits remain hidden. During search, the evaluator controls the training recipe (step budget, peak learning rate, schedule shape, decay fraction, and learning rate floor). 

\subsection{Hardware Information}
\label{app:hardware}

Table~\ref{tab:gpu-specs} shows more information on the GPUs used.

\begin{table}[t]
\centering
\small
\setlength{\tabcolsep}{4pt}
\caption{\textbf{GPU specifications.} Bandwidth is per device peak; Precision lists selected native matrix multiply formats.}
\label{tab:gpu-specs}
\begin{tabular}{@{}llrrl@{}}
\toprule
GPU class & Memory & Capacity (GB) & Bandwidth (TB/s) & Precision \\
\midrule
Datacenter & HBM3  & 80 & 3.35 & FP16, BF16, FP8 \\
Datacenter & HBM2e & 80 & 2.04 & FP16, BF16 \\
Consumer   & GDDR7 & 32 & 1.79 & FP16, BF16, FP8, MXFP8, FP4 \\
\bottomrule
\end{tabular}
\end{table}

\subsection{Additional Agent Discovered Methods}
\label{app:novel}

We concisely summarize two additional mathematical methods discovered in Scenario~B that complement the three examples in \cref{sec:novel}.

\noindent\textbf{Consistent $\Delta$ 8-bit FlashAttention.}
To run varlen attention on 8-bit tensor cores, the agent combined three algebraic identities in custom Triton kernels.
For forward $QK^\top$, per channel smoothing $\lambda_c = \sqrt{\max_t \lvert K_{tc}\rvert / \max_t \lvert Q_{tc}\rvert}$ balances outliers while preserving $(Q\operatorname{diag}(\lambda))(K\operatorname{diag}(\lambda)^{-1})^\top = QK^\top$.
In backward $dS = P \odot (dP - \Delta)$, where cancellation amplifies mismatch between quantized $dP = dO\,V^\top$ and $\Delta$, the agent computes $\Delta_i = \sum_c \hat{dO}_{ic} O_{ic} / \mu_c$ directly from the quantized $\hat{dO}$ and forward output $O$ produced with the same quantized $\hat{V}$ ($\mu_c = \sqrt{\max_t \lvert V_{tc}\rvert}$).
For $dQ = dS\,K$, keys are mean centered as $K - \bar{K}$ before int8 quantization, exact because $\sum_j dS_{ij} = 0$, cutting runtime by 4.0\,\% overall and 11\,\% on long sequences.

\noindent\textbf{Active vocabulary head updates and input gradient asymmetry.}
Over a 152k vocabulary, dense head updates are costly while freezing the head harms adaptation to new tokens.
Per token tile, the agent collects active rows $\mathcal{V}_{\mathrm{act}} = \{v : \max_t \lvert p_t(v) - \mathbf{1}_{v = y_t}\rvert \ge 10^{-4}\}$, which comprise 1.3--3.4\,\% of rows yet carry nearly all gradient norm, and updates $dW_v$ on device without host synchronization, cutting feedback runtime by 4.1\,\%.
Applying the same mask to input gradients $dH = (P - Y)W$ degraded local validation loss despite 0.02--0.25\,\% aggregate norm error: confident tokens fall below $10^{-4}$, leaving a coherent bias $\approx -\varepsilon W_{y_t}$, so the agent kept $dH$ dense.

\end{document}